\documentclass[english,preprint]{lni}
\usepackage{tabularx}
\usepackage{csquotes}
\usepackage{subcaption}

\usepackage{tikz}
\usepackage{pgfplots}
\pgfplotsset{compat=1.18}
\usepackage{pgfplotstable}
\usepackage{booktabs} 
\usepackage{fancyhdr}
\begin{document}
\pagestyle{fancy}

\title[AI-Assisted Programming for EUD]{Feasibility of AI-Assisted Programming for End-User Development}
\subtitle{Preprint \\ \footnotesize{Accepted for NLPIR 2025}}
 \author[1]{Irene Weber}{irene.weber@hs-kempten.de}{0000-0003-2743-1698}%
 \affil[1]{University of Applied Sciences Kempten\\Faculty of Mechanical Engineering\\Bahnhofstraße 61\\87435 Kempten\\Germany}

\maketitle

\begin{abstract}
End-user development,where non-programmers create or adapt  their own digital tools, can play a key role in driving digital transformation within organizations. Currently, low-code/no-code  platforms are widely used to enable end-user development through visual programming, minimizing the need for manual coding.

Recent advancements in generative AI, particularly large language model-based assistants and “copilots”, open new possibilities, as they may enable end users to generate and refine programming code and build apps directly from natural language prompts. This approach, here referred to as  AI-assisted end-user coding, promises greater flexibility, broader applicability, faster development, improved reusability, and reduced vendor lock-in compared to the established visual  LCNC platforms.

This paper investigates whether AI-assisted end-user coding  is a feasible paradigm for end-user development, which may complement or even replace the LCNC model in the future. To explore this, we conducted a case study in which non-programmers were asked to develop a basic web app   through interaction with AI assistants.The majority of study participants successfully completed the task in reasonable time and also expressed support for  AI-assisted end-user coding as a viable approach for end-user development. The paper presents the study design, analyzes the outcomes, and discusses  potential implications for practice, future research, and academic teaching.
\end{abstract}


\subsection*{}
{\small
This preprint has
not undergone peer review or any post-submission improvements or
corrections. The Version of Record of this contribution will be published in \emph{Herwig Unger, Phayung Meesad  (eds.): Advances in Natural Language
Processing and Information Retrieval}}

\pagebreak

\section{Introduction}\label{sec:intro}

Companies are driving digitalization to gain efficiency, reduce error rates, avoid media disruptions, and collect high-quality, structured data  from the start of all processes. However, capacity bottlenecks in IT departments often slow down the development and rollout of digital solutions. End-User Development (EUD) and Business-Managed IT (BMIT) are two related approaches that help alleviate such constraints. In EUD, end users develop or modify their own applications, such as workflow automations or digital tools, tailored to their tasks \cite{liebermanEndUserDevelopment2006, paternoNewPerspectivesEndUser2017}. BMIT refers to situations where business units, rather than central IT departments, manage the development or delivery of IT systems \cite{kopperBusinessmanagedITConceptual2018}.

EUD and BMIT have been found to offer various benefits. Solutions stemming from these approaches tend to be well aligned with user needs and can address niche problems that central IT might overlook or deprioritize due to resource constraints. EUD and BMIT can also enhance a company’s adaptability and capacity for innovation. Additionally, they contribute to overall job satisfaction, as employees often experience greater motivation and autonomy through their direct involvement and feel a stronger sense of ownership over the tools they create and use. \cite{kopperBusinessmanagedITConceptual2018}

Often, companies enable EUD by providing visual low-code/no-code (LCNC) platforms. End users can create digital tools—such as automation flows, data entry forms, app controls, or data displays—by selecting, connecting, and configuring graphical components. In addition to supporting visual development, these LCNC platforms typically include connectors to business applications and host the developed solutions.

Modern artificial intelligence (AI) technologies, especially large language models (LLM), are expected to have immense potential for  digitalization in businesses. These LLMs are pre-trained on massive amounts of text data (including computer programs and code) and later f ine-tuned to follow instructions, use tools, and more. Triggered by language inputs termed prompts, they can perform a wide range of tasks and exhibit intelligence that seems close to human-level.
Two emerging approaches in this realm are the AI-copilot paradigm, where LLMs solve tasks in cooperation with humans, and the AI-agent paradigm, where LLMs are trusted to solve tasks more or less autonomously.

Another approach is to exploit the coding capabilities of LLMs. Given a textual description of the problem to be solved and, potentially, the system context in which the software should run, these models can automatically generate programming code and produce working software solutions in many programming languages. 
With AI now making it easier to write code, a new form of end-user development (EUD) may become feasible: employees building digital tools by writing code with  LLMs. The approach of LLM-assisted end-user app programming, here referred to as  end-user coding (EUCod)  may help overcome some of the limitations associated with  LCNC platforms,  as recently discussed in tech-oriented media~\cite{tozzi2024comparison, vanHemert2025nocode, waehner2025generative}: 

\begin{enumerate} 
\item \textbf{Better Customization}:
Coding allows to tailor software precisely to  individual  needs, surpassing  limitations and pre-configurations of standardized LCNC solutions.

\item \textbf{Higher User Acceptance Through a Unified “One-Stop-Shop” Solution}:
Users no longer need to rely on multiple specialized tools (e.g., platforms for building data collection forms, poll-based meeting planners etc.). Instead, they can work within a single environment, avoiding the hassle of remembering,  managing and switching between different tools, each with its own learning curve.

\item \textbf{Faster development}:
Compared to the numerous manual interactions with graphical components required when working with  visual LCNC platforms, instructing  an LLM to write or adapt code can be faster.

\item \textbf{Improved accessibility and reusability; elimination of vendor lock-in}:
Unlike solutions built on proprietary LCNC platforms, self-generated custom and self-used code\footnote{However, there is a risk that code generated by LLMs   infringes on third-party copyrights.} is in the ownership of the coder, fully accessible and easier to modify, extend, reuse, and transfer to other environments.

\end{enumerate}

This paper presents an  experimental study investigating the viability of LLM-assisted EUCod  as a potentially emerging approach to EUD. 
Specifically, we  to explore  whether non-programmers are  able and willing to develop basic digital tools with the assistence of  LLMs. To this aim, we  asked study participants to build a simple web application, specifically, a survey form, using ChatGPT or a similar LLM-based chat as their development assistant.
The paper is structured as follows: Section~2  provides an overview of  related work as well as approaches encountered in practice. Section~3 presents  the experimental setup and report the results. Section~4 discusses the findings and their implications and concludes.

\section{Related Work}\label{sec:related}

Various approaches apply large language models (LLMs) to enable end users, particularly those without formal training in software development, to create application code\footnote{The author does not suggest that these approaches guarantee success or are free of risk. AI-generated code may contain errors, inefficiencies, or security vulnerabilities, and therefore requires careful testing and oversight}. A recent literature review on the role of end users in LLM-assisted end-user development is \cite{gargioniExploringRoleEnd2024}.

A growing body of work investigates the use of \textbf{LLMs for robot programming}. Several studies explicitly address robot programming by end users with LLM assistance~\cite{ge2024cocobo, karliAlchemistLLMAidedEndUser2024, bimbattiCanChatGPTSupport2023, gargioniPreparationPersonalizedMedicines2025}. These approaches typically provide integrated development environments (IDEs) that feature a chat interface where users describe requirements in natural language. In addition, they offer visual representations of the LLM-generated robot control code, such as flowcharts or block diagrams, which users can inspect and modify.
Some of these works also describe the predefined instructions (system prompts) used to inform the LLM about available robot actions and the expected output formats~\cite{ge2024cocobo, karliAlchemistLLMAidedEndUser2024, gargioniPreparationPersonalizedMedicines2025}.

A further strand of research investigates \textbf{LLMs as educational tools for novice programmers}~\cite{feldmanNonExpertProgrammersGenerative2024, pratherWideningGapBenefits2024, nguyenHowBeginningProgrammers2024, chenLearningAgentbasedModeling2024}. For example, \cite{feldmanNonExpertProgrammersGenerative2024} examine how undergraduate students use LLMs to complete beginner-level programming tasks, deriving insights for future  programming education.

Some authors \textbf{combine AI-assisted  EUCod with LCNC platforms}, such as 
Oracle APEX \cite{gorissenEndUserDevelopmentOracle2024}  or the Google Home companion app. Oracle APEX is a LCNC development platform for building web applications on top of Oracle databases \cite{oracleOracleAPEX}. 
In the approach proposed by~\cite{gorissenEndUserDevelopmentOracle2024}, end users can modify existing APEX applications using natural language requests. An LLM translates these requests into calls to APEX’s Application Editing API.
\cite{barricelliPosterNaturalLanguage2023} explore speech-only conversational approaches to creating routines for virtual assistants on smart speakers, using the Google Home companion app as a model.

\cite{huangCreatingUserInterface2021, jiangGenLineGenFormTwo2021} are early studies on \textbf{LLM-supported end-user web and app development}, exploring code generation for websites with the help of language models. \cite{kimStyletteStylingWeb2022} presents a browser extension that supports CSS styling through LLM assistance, all prior to the advent of ChatGPT in November 2022.
More recent work includes a proof-of-concept by \cite{caloLeveragingLargeLanguage2023}, which generates HTML and CSS code for websites based on natural language descriptions. Their system uses pre-engineered prompts to guide the LLM in producing outputs that conform to a specific template, making the responses directly parsable. The authors plan to integrate this approach into LCNC platforms to improve its accessibility and effectiveness for non-technical users.

Recently, \textbf{commercial AI-based platforms for web and app development} have emerged that aim to empower end users to create applications through natural language inputs without writing code~\cite{kortixaiSoftgenAIWeb2024, lovablelabsincorporatedLovable2025, stackblitzBoltnew2025}. Lovable.dev~\cite{lovablelabsincorporatedLovable2025} and SoftGen.ai~\cite{kortixaiSoftgenAIWeb2024} support the generation of full-stack web applications, while Bolt.new~\cite{stackblitzBoltnew2025} additionally enables the creation of mobile apps. These platforms offer third-party integrations with services such as databases, payment systems, and more. Notably, Lovable.dev is the commercial successor to GPT-Engineer, an open-source command-line tool for AI-driven code generation and execution based on natural language software specifications, available on GitHub~\cite{osikaGptengineer2023}.

\textbf{Vibe coding} is a recently coined term for AI-driven programming, in which developers iteratively prompt LLMs to generate and refine code with minimal manual intervention~\cite{karpathyTheresNewKind2025}. So far, the approach seems  primarily been promoted as a productivity aid for skilled programmers, rather than as a tool for non-programmers. A first online course introducing vibe coding is  now available~\cite{deeplearning.aiVibeCoding1012025}.

\section{Empirical Study on AI-Assisted End Users Coding}

We conducted an experimental study to evaluate whether EUCod may be a feasible approach to EUD, possibly  replacing the visual development paradigm  supported by LCNC platforms in the future.

The study was conducted at a German university of applied sciences with  students of industrial engineering as  {\bf participants}, including both bachelor’s and master’s degree students. These students commonly have  basic coding skills from introductory programming classes, but  no further formal education as software or app developers.  

As the task of designing a survey  is likely to arise  in roles such as knowledge work, project management, and related fields, participants were asked to implement a survey app as the \textbf{study task}. To avoid the need for a complex development environment or deployment process, the technical setup was deliberately kept simple: the form was to be created as a single HTML file, complete with CSS styling and JavaScript for interactive elements, saved locally, and previewed in a standard web browser. VS Code was recommended as the code editor for development. For collecting survey responses, participants were instructed to connect to a Google Sheet  via a Google Apps Script, providing a lightweight and accessible backend solution.

{\bf Task Instructions} specified further details as follows:
Participants were instructed to design a meaningful survey on a topic of their choice. The survey should include a variety of input elements (e.g., both checkboxes and radio buttons), but avoid overly complex features. If possible, the form was to include a dynamic element,  such as a follow-up question revealed conditionally based on user input.

Participants were instructed to begin by describing their task to the LLM and then develop the form iteratively by refining and expanding it through further prompts. To help the participants  get started, a basic, non-mandatory starting prompt was suggested: \textit{“How can I design an HTML survey form that\ldots?”}
They were encouraged to ask the LLM for explanations in case they encountered challenges or unclear concepts.

{\bf The experimental setup} was designed  to foster a relaxed, workshop-like atmosphere inviting to exploration. 
Participants were given the opportunity to work on the task during lecture hours using university lab computers. Master’s students were explicitly instructed to work in pairs, while bachelor’s students received no specific collaboration instructions. However, this distinction was not strictly enforced, and some variation in group work occurred. Use of university lab computers was also not enforced; students were free to work on their own devices or at home. 
Master’s students worked on the task as part of a graded assignment, whereas bachelor’s students could engage with it as an optional, ungraded lab exercise. 
At the beginning of the study, a pre-developed example survey form was demonstrated, and its components and technical setup were explained during an approximately 20-minute introduction.

Participants progressed far more quickly than expected and were asked to complete a {\bf follow-up survey}  on their project status and experiences the following week. 
The original task instructions, the survey form and the collected responses (in German)  are available in the  {\bf online appendix}\footnote{\url{https://weberi.github.io/2025_AI-Assisted-End-User-Coding}
}.

\subsection{Results}

\begin{figure}
\centering 
\begin{subfigure}{0.20\textwidth}\scriptsize
  \centering
  \begin{tikzpicture}
\begin{axis}[
    ybar stacked,
    height=5.5cm,
    axis line style={draw=white},
    axis y line=none,    
    xlabel={},   
    ylabel={},      
    ytick=\empty,   
    tick style={draw=none},
    x=21pt,     
    bar width=12pt,
    enlargelimits=0.45,
    symbolic x coords={Seldom, Oft.,  Very Oft.},
    xtick=data,
    xticklabel style={font=\scriptsize},
    enlarge x limits=0.25,
    ymin=0,
    ymax=25,
    enlarge y limits={value=0,upper},
    nodes near coords,
    nodes near coords align={center},
]
\addplot+[
    fill=gray!25, 
    draw=black,
    nodes near coords, 
    nodes near coords style={black}
] 
plot coordinates {(Seldom,3) (Oft.,2) (Very Oft.,7)  };

\addplot+[
    fill=gray!60, 
    draw=black, 
    nodes near coords, 
    nodes near coords style={black}
] 
plot coordinates {(Seldom,1) (Oft.,11)  (Very Oft.,9) };
\end{axis}
  \end{tikzpicture}
  \caption{AI Usage}
\end{subfigure}\hfill \begin{subfigure}{0.20\textwidth} \scriptsize
  \centering
  \begin{tikzpicture}
\begin{axis}[
    ybar stacked,
    height=5.5cm,
    axis line style={draw=white},
    axis y line=none,    
    xlabel={},   
    ylabel={},      
    ytick=\empty,   
    tick style={draw=none},
    x=21pt,     
    bar width=12pt,
    enlargelimits=0.45,
    symbolic x coords={Low,Medium, High},
    xtick=data,
    xticklabel style={font=\scriptsize},
    enlarge x limits=0.25,
    ymin=0,
    ymax=25,
    enlarge y limits={value=0,upper},
    nodes near coords,
    nodes near coords align={center}
]
\addplot+[
    fill=gray!25, 
    draw=black,
    nodes near coords, 
    nodes near coords style={black}
] 
plot coordinates { (Low,5) (Medium,4 )  (High,3)};

\addplot+[
    fill=gray!60, 
    draw=black, 
    nodes near coords, 
    nodes near coords style={black}
] 
plot coordinates {(Low,5)  (Medium,14 )  (High,2)};
\end{axis}
  \end{tikzpicture}
  \caption{AI Competence}
\end{subfigure}\hfill \begin{subfigure}{0.2\textwidth} \scriptsize
  \centering
  \begin{tikzpicture}
\begin{axis}[
    ybar stacked,
    height=5.5cm,
    axis line style={draw=white},
    axis y line=none,    
    xlabel={},   
    ylabel={},      
    ytick=\empty,   
    tick style={draw=none},
    x=21pt,     
    bar width=12pt,
    enlargelimits=0.45,
    symbolic x coords={None, Some, Good},
    xtick=data,
    xticklabel style={font=\scriptsize},
    enlarge x limits=0.25,
    ymin=0,
    ymax=25,
    enlarge y limits={value=0,upper},
    nodes near coords,
    nodes near coords align={center}
]
\addplot+[
    fill=gray!25, 
    draw=black,
    nodes near coords, 
    nodes near coords style={black}
] 
plot coordinates {(None,3) (Some,7 ) (Good,2)  };

\addplot+[
    fill=gray!60, 
    draw=black, 
    nodes near coords, 
    nodes near coords style={black}
] 
plot coordinates {(None,6) (Some,13 ) (Good,2)};
\end{axis}
  \end{tikzpicture}
  \caption{Coding}
\end{subfigure}\hfill \begin{subfigure}{0.2\textwidth} \scriptsize
  \centering
  \begin{tikzpicture}

\begin{axis}[
    ybar stacked,
    height=5.5cm,
    axis line style={draw=white},
    axis y line=none,    
    xlabel={},   
    ylabel={},      
    ytick=\empty,   
    tick style={draw=none},
    x=21pt,     
    bar width=12pt,
    enlargelimits=0.45,
    symbolic x coords={None, Some, Good},
    xtick=data,
    xticklabel style={font=\scriptsize},
    enlarge x limits=0.25,
    ymin=0,
    ymax=25,
    enlarge y limits={value=0,upper},
    nodes near coords,
    nodes near coords align={center}
]
\addplot+[
    fill=gray!25, 
    draw=black,
    nodes near coords, 
    nodes near coords style={black}
] 
plot coordinates {(None,8) (Some,4 ) (Good,0)  };

\addplot+[
    fill=gray!60, 
    draw=black,
    nodes near coords, 
    nodes near coords style={black}
] 
plot coordinates {(None,13) (Some,8 ) (Good,0)};
\end{axis}

  \end{tikzpicture}
  \caption{HTML}
\end{subfigure}\hfill \begin{subfigure}{0.2\textwidth} \scriptsize
  \centering
  \begin{tikzpicture}
\begin{axis}[
    ybar stacked,
    height=5.5cm,
    axis line style={draw=white},
    axis y line=none,    
    xlabel={},   
    ylabel={},      
    ytick=\empty,   
    tick style={draw=none},
    x=21pt,     
    bar width=12pt,
    enlargelimits=0.45,
    symbolic x coords={None, Some, Good},
    xtick=data,
    xticklabel style={font=\scriptsize},
    enlarge x limits=0.25,
    ymin=0,
    ymax=25,
    enlarge y limits={value=0,upper},
    nodes near coords,
    nodes near coords align={center}
]
\addplot+[
    fill=gray!25, 
    draw=black,
    nodes near coords, 
    nodes near coords style={black}
] 
plot coordinates {(None,10) (Some,2) (Good,0)};

\addplot+[
    fill=gray!60, 
    draw=black,
    nodes near coords, 
    nodes near coords style={black}
] 
plot coordinates {(None,11) (Some,10)  (Good,0)};
\end{axis}
  \end{tikzpicture}
  \caption{Web Development}
\end{subfigure}
\caption{Self assessment of prior experience. 
\protect\tikz[baseline=0ex]{\protect\draw[fill=gray!25, draw=black] (0,0) rectangle (0.3,0.2);} BA, 
\protect\tikz[baseline=-0ex]{\protect\draw[fill=gray!60, draw=black] (0,0) rectangle (0.3,0.2);} MA.}\label{fig:prior}
\end{figure}
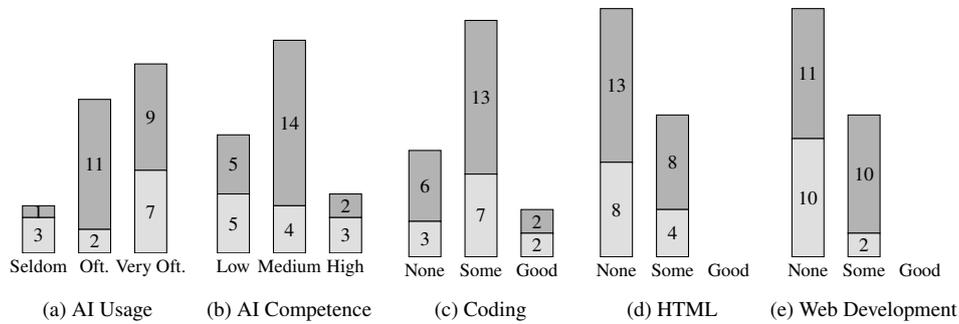

The survey received 33 usable responses, excluding one incomplete submission and those that were submitted after the deadline. Of these, 21 came from the master’s group and 12 from the bachelor’s group. All participants completed the survey individually, not as teams.
The survey included questions in which participants self-assessed their frequency of AI use and their level of expertise in relevant skills, see~\cref{fig:prior}.

Most participants reported using AI tools \emph{often} or \emph{very often}. Responses for AI Competence are skewed toward \emph{medium} competence, with few rating themselves as \emph{high}. 
 Participants reported \emph{some} coding experience more frequently than \emph{none} or \emph{good}. Only few considered their coding skills \emph{good}. 
Participants indicated even less familiarity with HTML and web development. The majority reported their  knowledge in these areas as \emph{none} , and no participants rated their skills as \emph{good}.

\begin{figure}
\centering 
\begin{subfigure}{0.23\textwidth}\scriptsize
  \centering
  \begin{tikzpicture}
\begin{axis}[
    ybar stacked,
    height=5.8cm,
    axis line style={draw=white},
    axis y line=none,    
    xlabel={},   
    ylabel={},      
    ytick=\empty,   
    tick style={draw=none},
    x=25pt,     
    bar width=12pt,
    enlargelimits=0.45,
    symbolic x coords={No, Yes},
    xtick=data,
    enlarge x limits=0.25,
    ymin=0,
    ymax=30,
    enlarge y limits={value=0,upper},
    nodes near coords,
    nodes near coords align={center},
]
\addplot+[
    fill=gray!25, 
    draw=black,
    nodes near coords, 
    nodes near coords style={black}
] 
coordinates {(No,6) (Yes,6)};

\addplot+[
    fill=gray!60, 
    draw=black, 
    nodes near coords, 
    nodes near coords style={black}
] 
coordinates {(No,3) (Yes,18)};
\end{axis}
  \end{tikzpicture}
  \caption{Task Success}
\end{subfigure}\hfill \begin{subfigure}{0.25\textwidth}\scriptsize
  \centering
  \begin{tikzpicture}
\begin{axis}[
    ybar stacked,
    height=5.8cm,
    axis line style={draw=white},
    axis y line=none,
    xlabel={},
    ylabel={},
    ytick=\empty,
    tick style={draw=none},
    x=30pt,
    bar width=12pt,
    enlargelimits=0.45,
    symbolic x coords={
        Easy,
        Reason.,
        Diff. 
    },
    xtick=data,
    enlarge x limits=0.25,
    ymin=0,
    ymax=28,
    enlarge y limits={value=0,upper},
    nodes near coords,
    nodes near coords align={center},
]
\addplot+[
    fill=gray!25,
    draw=black,
    nodes near coords,
    nodes near coords style={black}
]
coordinates {
    (Easy,1)
    (Reason.,8)
    (Diff.,3)
};

\addplot+[
    fill=gray!60,
    draw=black,
    nodes near coords,
    nodes near coords style={black}
]
coordinates {
    (Easy,2)
    (Reason.,16)
    (Diff.,3)
};
\end{axis}
  \end{tikzpicture}
  \caption{Perceived Effort}
\end{subfigure}\hfill\begin{subfigure}{0.23\textwidth}\scriptsize
  \centering
  \begin{tikzpicture}
\begin{axis}[
    ybar stacked,
    height=5.8cm,
    axis line style={draw=white},
    axis y line=none,
    xlabel={},
    ylabel={},
    ytick=\empty,
    tick style={draw=none},
    x=28pt,
    bar width=12pt,
    enlargelimits=0.45,
    symbolic x coords={
        No,
        Yes
    },
    xtick=data,
    enlarge x limits=0.25,
    ymin=0,
    ymax=30,
    enlarge y limits={value=0,upper},
    nodes near coords,
    nodes near coords align={center},
]
\addplot+[
    fill=gray!25,
    draw=black,
    nodes near coords,
    nodes near coords style={black}
]
coordinates {
    (No,2)
    (Yes,10)
};

\addplot+[
    fill=gray!60,
    draw=black,
    nodes near coords,
    nodes near coords style={black}
]
coordinates {
    (No,3)
    (Yes,18)
};
\end{axis}
  \end{tikzpicture}
  \caption{Recommend}
\end{subfigure} \hfill\begin{subfigure}{0.23\textwidth}\scriptsize
  \centering
  \begin{tikzpicture}
\begin{axis}[
    ybar stacked,
    height=5.8cm,
    axis line style={draw=white},
    axis y line=none,
    xlabel={},
    ylabel={},
    ytick=\empty,
    tick style={draw=none},
    x=25pt,
    bar width=12pt,
    enlargelimits=0.45,
    symbolic x coords={
        No,
        Maybe,
        Yes
    },
    xtick=data,
    enlarge x limits=0.25,
    ymin=0,
    ymax=30,
    enlarge y limits={value=0,upper},
    nodes near coords,
    nodes near coords align={center},
]
\addplot+[
    fill=gray!25,
    draw=black,
    nodes near coords,
    nodes near coords style={black}
]
coordinates {
    (No,2)
    (Maybe,3)
    (Yes,7)
};

\addplot+[
    fill=gray!60,
    draw=black,
    nodes near coords,
    nodes near coords style={black}
]
coordinates {
    (No,0) 
    (Maybe,12)
    (Yes,9)
};
\end{axis}
  \end{tikzpicture}
  \caption{Use}
\end{subfigure}
\caption{Survey responses on AI-assisted end-user coding including task success, perceived effort, and openness to recommending and using the approach in enterprise settings 
\protect\tikz[baseline=0ex]{\protect\draw[fill=gray!25, draw=black] (0,0) rectangle (0.3,0.2);} BA, 
\protect\tikz[baseline=-0ex]{\protect\draw[fill=gray!60, draw=black] (0,0) rectangle (0.3,0.2);} MA.}\label{fig:outcome}
\end{figure}
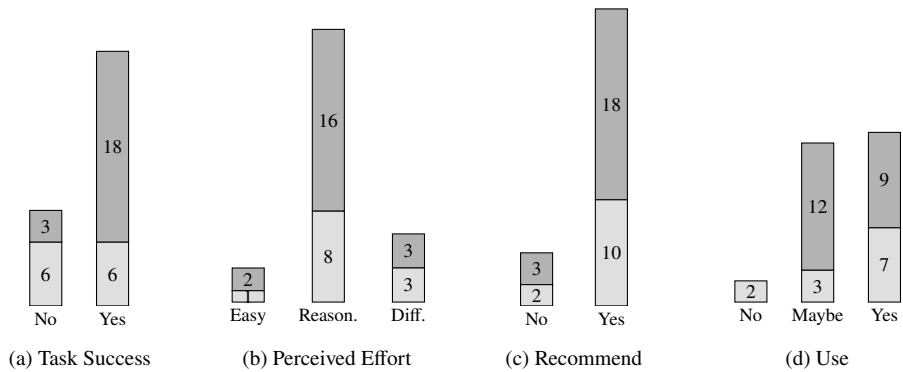

\cref{fig:outcome} summarizes the main outcomes of the survey. 
Most participants (24 out of 33) reported successful completion of the task, meaning that they had developed a functional survey. Success was notably higher among master’s students, 85\% of whom completed the task successfully, compared to 50\% among bachelor’s students.

Regarding the perceived effort,  a clear  majority comprising 24 participants   rated the effort-to-outcome ratio as \emph{appropriate}. Three  participants had found the task \emph{easier} than expected,  while six participants described the process as \emph{difficult}. 

Participants were also asked whether they would recommend offering the AI-assisted end-user coding approach within enterprise settings. The overwhelming majority (28 out of 33) responded \emph{yes}. When asked whether they would personally use the approach in the future, 16 participants answered \emph{yes}, and 15 selected \emph{maybe}. Only two bachelor’s students responded with a clear \emph{no}.

\begin{figure}
\hfill
\begin{subfigure}{0.3\textwidth}\scriptsize
  \centering
  \begin{tikzpicture}
\begin{axis}[
    ybar stacked,
    height=4cm,
    axis line style={draw=white},
    axis y line=none,
    xlabel={},
    ylabel={},
    ytick=\empty,
    tick style={draw=none},
    x=24pt,
    bar width=12pt,
    enlargelimits=0.45,
    symbolic x coords={
        <1,
        2--3,
        4--5,
        >6
    },
    xtick=data,
    enlarge x limits=0.25,
    ymin=0,
    ymax=18,
    enlarge y limits={value=0,upper},
    nodes near coords,
    nodes near coords align={center},
]
\addplot+[
    fill=gray!25,
    draw=black,
    nodes near coords,
    nodes near coords style={black}
]
coordinates {
    (<1,1)
    (2--3,9)
    (4--5,1)
    (>6,1)
};

\addplot+[
    fill=gray!60,
    draw=black,
    nodes near coords,
    nodes near coords style={black}
]
coordinates {
    (<1,0)
    (2--3,3)
    (4--5,9)
    (>6,9)
};
\end{axis}
  \end{tikzpicture}
  \caption{Hours Spent by participants \\ \protect\tikz[baseline=0ex]{\protect\draw[fill=gray!25, draw=black] (0,0) rectangle (0.3,0.2);}. BA, 
\protect\tikz[baseline=-0ex]{\protect\draw[fill=gray!60, draw=black] (0,0) rectangle (0.3,0.2);} MA.}\label{fig:timea}
\end{subfigure}\hfill\begin{subfigure}{0.3\textwidth}\scriptsize
  \centering
  \begin{tikzpicture}
\begin{axis}[
    ybar stacked,
    height=4cm,
    axis line style={draw=white},
    axis y line=none,
    xlabel={},
    ylabel={},
    ytick=\empty,
    tick style={draw=none},
    x=24pt,
    bar width=12pt,
    enlargelimits=0.45,
    symbolic x coords={
        <1,
        2--3,
        4--5,
        >6
    },
    xtick=data,
    enlarge x limits=0.25,
    ymin=0,
    ymax=18,
    enlarge y limits={value=0,upper},
    nodes near coords,
    nodes near coords align={center},
]
\addplot+[
    fill=gray!10,
    draw=black,
    nodes near coords,
    nodes near coords style={black}
]
coordinates {
    (<1,0)
    (2--3,9)
    (4--5,8)
    (>6,7)
};

\addplot+[
    fill=gray!90,
    draw=black,
    nodes near coords,
    nodes near coords style={white}
]
coordinates {
    (<1,1)
    (2--3,3)
    (4--5,2)
    (>6,3)
};
\end{axis}
  \end{tikzpicture}
  \caption{Hours Spent vs.\ Task Success. \\ \protect\tikz[baseline=0ex]{\protect\draw[fill=gray!10, draw=black] (0,0) rectangle (0.3,0.2);} Task Success, 
\protect\tikz[baseline=-0ex]{\protect\draw[fill=gray!90, draw=black] (0,0) rectangle (0.3,0.2);} No Success.}\label{fig:timesuccess}
\end{subfigure}\hfill
\begin{subfigure}{0.3\textwidth}\scriptsize
  \centering
  \begin{tikzpicture}
\begin{axis}[
    ybar stacked,
    height=4cm,
    axis line style={draw=white},
    axis y line=none,
    xlabel={},
    ylabel={},
    ytick=\empty,
    tick style={draw=none},
    x=24pt,
    bar width=12pt,
    enlargelimits=0.45,
    symbolic x coords={
        <1,
        2--3,
        4--5,
        >6
    },
    xtick=data,
    enlarge x limits=0.25,
    ymin=0,
    ymax=14,
    enlarge y limits={value=0,upper},
    nodes near coords,
    nodes near coords align={center},
]
\addplot+[
    fill=gray!10,
    draw=black,
    nodes near coords,
    nodes near coords style={black}
]
coordinates {
    (<1,1)
    (2--3,8)
    (4--5,3)
    (>6,4)
};

\addplot+[
    fill=gray!90,
    draw=black,
    nodes near coords,
    nodes near coords style={white}
]
coordinates {
    (<1,0)
    (2--3,4)
    (4--5,7)
    (>6,6)
};
\end{axis}
  \end{tikzpicture}
  \caption{Hours Spent vs.\ Survey Complexity. \protect\tikz[baseline=0ex]{\protect\draw[fill=gray!10, draw=black] (0,0) rectangle (0.3,0.2);} Simple, 
\protect\tikz[baseline=-0ex]{\protect\draw[fill=gray!90, draw=black] (0,0) rectangle (0.3,0.2);} Complex.}\label{fig:timecomplex}
\end{subfigure}
\hfill
\caption{Time spent on survey development.}\label{fig:timespent}
\end{figure}
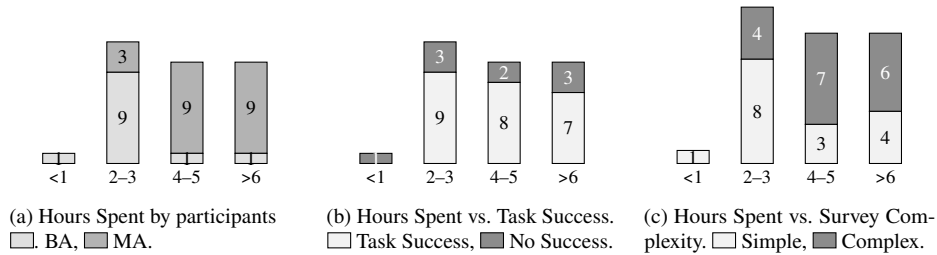

The survey also asked participants to report the time spent developing the survey app, see~\cref{fig:timesuccess}. Successfully completed app developments  are distributed across all time categories. The majority of participants who succeeded had spent at least 2--3 hours on the task, suggesting that a moderate time investment was generally sufficient for success.
Participants also indicated the complexity of their surveys.  A survey was considered complex if it included dynamic elements that adapt to user input, implemented input validation, or similar features. Nearly half of the participants rated their surveys as complex. Notably, complexity did not  clearly correlate with longer development times, see ~\cref{fig:timecomplex}.

A further group of survey questions addressed  difficulties encountered during  app development, organizational measures enterprises should take when introducing AI-assisted EUCod, and the AI models used for the task. These questions allowed unrestricted responses. The free-text answers were analyzed and categorized by the author of this paper. Some responses addressed multiple themes and were therefore assigned to more than one category. See ~\cref{tab:free-texts} for the results.

\begin{table}[ht]
\centering
\begin{subtable}[t]{0.45\textwidth}
\centering
\caption{Reported Difficulties}
\begin{tabular}{@{}lr@{}}
\toprule
\textbf{Issue} & \textbf{Count} \\
\midrule
Analysis (GS)                        & 3  \\
Prompting                         & 6  \\
HTML \& Web Dev           & 7  \\
GS Integration                         & 15 \\
	Faulty Codee       & 4  \\
Model restrictions            & 1  \\
\bottomrule
\end{tabular}
\end{subtable}
\hfill
\begin{subtable}[t]{0.23\textwidth}
\centering
\caption{Needs}
\begin{tabular}{@{}lr@{}}
\toprule
\textbf{Provision} & \textbf{Count} \\
\midrule
Training              & 18 \\
Platform  & 10 \\
Support     & 4  \\
\bottomrule
\end{tabular}
\end{subtable}
\hfill
\begin{subtable}[t]{0.23\textwidth}
\centering
\caption{AI assistants}
\begin{tabular}{@{}lr@{}}
\toprule
\textbf{Assistant} & \textbf{Count} \\
\midrule
ChatGPT     & 31 \\
DeepSeek    & 8  \\
Copilot     & 2  \\
Gemini      & 1  \\
Perplexity  & 3  \\
Lovable     & 2  \\
\bottomrule
\end{tabular}
\end{subtable}

\caption{Overview of reported development challenges, support needs, and AI assistants used.}
\label{tab:free-texts}
\end{table}

The most frequent development issue concerned the integration with Google Sheets, followed by challenges related to HTML and web development, and formulating effective prompts. Less commonly mentioned were problems with faulty code generation, and technical model restrictions, such as context length or usage limits.
Some participants interpreted the study task as including the analysis of survey responses within Google Sheets, which they also found to be challenging.
When asked what support enterprises should provide to enable AI-assisted end-user development, most participants emphasized the need for introductory training, followed by the demand for provision of a dedicated, easy-to-use development platform. A few additionally mentioned access to human experts or a user community for support.
Among the AI assistants used, ChatGPT clearly dominated, but several participants also reported using alternative tools or combinations of tools.

\section{Discussion and Conclusion}

This study explores the feasibility of AI-assisted application programming (EUCod) for EUD. Participants with minimal formal IT education or software development experience were tasked with developing a survey application using AI assistance. The application featured a frontend built with HTML, CSS, and JavaScript, and a backend implemented via Google Sheets and Google Apps Script for storing responses, qualifying it as a minimal full-stack application.

The majority of participants found the effort-to-outcome ratio appropriate and recommended that enterprises support AI-assisted EUCod for EUD initiatives. Nearly all participants expressed willingness to adopt this approach in the future. These findings suggest that AI-assisted EUCod could be a viable approach for EUD in organizations, especially as we can expect that  LLM assistants and also  AI competence among users will continue to improve over time.

While most participants succeeded in developing a functional survey app and connecting it to a simple backend within a reasonable timeframe, they also expressed a need for training, support, and guidance. Approximately 30\% suggested that enterprises should implement measures to facilitate EUCod, which the author interpreted as a demand for a dedicated development platform.
Offering integrations with backend services, ideally tailored to an organization’s existing system landscape,  a dedicated development platform could help alleviate challenges such as connecting with a data store, which was the main difficulty reported in our study. While introducing such a platform for AI-assisted EUCod may come with some of the known downsides of LCNC tools (see \cref{sec:intro}),  the advantages would probably outweigh them, as the platform would also  address other concerns mentioned in the survey, such as the need for clear guidelines, compliance, IT governance, and data security.

From the study designer’s perspective, participants completed the assigned task more quickly and effectively than expected. Nearly half of them designed “complex” surveys. The deliberate selection and experimentation with different AI assistants, along with the general openness toward AI-assisted EUCod, indicate a promising level of AI literacy among participants. Notably, many students, who will soon join the active workforce,  are already regular users of AI assistants.

As with any empirical study, several limitations must be acknowledged. First, the study was conducted with students of industrial engineering at a German university of applied sciences. While these participants had limited programming experience, they still possess a technical background and may not fully represent the diversity of typical end users in enterprise contexts. This affects the external validity of the findings.

Second, the study involved two distinguishable groups of participants who shared similar academic backgrounds but differed in experience and study conditions. Participation was voluntary for some and mandatory for others, which may have influenced their motivation and level of engagement. This variation, along with a relatively uncontrolled environment (e.g., some participants worked from home or in teams), introduced variability in how the task was approached. On the other hand, this diversity may also reduce the influence of the specific university setting and thus improve the generalizability of the results.

Third, the findings largely rely on participants’ self-assessments and on the categorization and interpretation of responses by a single researcher. Third, the findings largely rely on participants’ self-assessments and on the categorization and interpretation of responses by a single researcher. While this may limit objectivity, it appears acceptable within the scope of a pioneering exploratory study.

Despite these limitations, the study offers valuable initial insights into the feasibility of AI-assisted end-user coding and highlights it as a promising and relevant topic not only for future research, but also for university teaching.

The combination of EUD platforms with AI-assisted end-user coding (EUCod) is of particular interest, as it can take several different forms. First, as mentioned in \cref{sec:related}, dedicated platforms for AI-assisted coding are emerging. A second variant can be found in related work: the idea of programming the platform itself through API calls  that are generated with the assistance of an LLM. 

A third option, which is not typically subsumed under the umbrella of end-user development, involves using LLMs to generate workflows (or similar digital artifacts) that can be deployed and executed on LCNC platforms. A preliminary evaluation of this direction is provided by \cite{fillConceptualModelingLarge2023}, who explored ChatGPT’s ability to generate BPMN diagrams in a custom simplified JSON format. Own informal testing indicated that LLMs, particularly ChatGPT, are capable of generating flows in JSON that can be imported into and executed by the LCNC platform Node-RED \cite{openjsfoundationLowcodeProgrammingEventdriven}, which is often used in the context of home automation, the Internet of Things, and industrial automation.
  
This approach of generating executable digital artifacts beyond conventional source code may gain practical significance, as it represents an additional pathway for AI-assisted end-user development.


\end{document}